\documentclass{ijcaArticle}
\usepackage[none]{hyphenat}
\ijcaVolume{99}
\ijcaNumber{17}
\ijcaYear{2014}
\ijcaMonth{August}

\begin{document}

\title{Investigation of a chaotic spiking neuron model} 

\author{
   \large M. Alhawarat \\[-3pt]
   \normalsize Department of Computer Science,  \\[-3pt]
    \normalsize Salman Bin Abdulaziz University, \\[-3pt]
    \normalsize Al Kharj, 11942, Kingdom of Saudi Arabia \\[-3pt]
    \normalsize	m.alhawarat$@$sau.edu.sa \\[-3pt]
  \and
   \large T. Olde Scheper, \\[-3pt]
   \normalsize Dept. of Computing and Comm. Tech.,  \\[-3pt]
    \normalsize Oxford Brookes University, \\[-3pt]
    \normalsize Wheatley, OXFORD OX33 1HX, UK \\[-3pt]
    \normalsize	tvolde-scheper$@$brookes.ac.uk \\[-3pt]
\and
   \large N. T. Crook \\[-3pt]
   \normalsize Head of Computing and Comm. Tech.,  \\[-3pt]
    \normalsize Oxford Brookes University, \\[-3pt]
    \normalsize Wheatley, Oxford OX33 1HX, UK \\[-3pt]
    \normalsize	ncrook$@$brookes.ac.uk \\[-3pt]
}

\terms{Chaotic Neural Model, Chaos}

\maketitle

\begin{abstract}
  Chaos provides many interesting properties that can be used to achieve computational tasks. Such properties are sensitivity to initial conditions, space filling, control and synchronization. Chaotic neural models have been devised to exploit such properties. In this paper, a chaotic spiking neuron model is investigated experimentally. This investigation is performed to understand the dynamic behaviours of the model.

  The aim of this research is to investigate the dynamics of the nonlinear dynamic state neuron (NDS) experimentally. The experimental approach has revealed some quantitative and qualitative properties of the NDS model such as the control mechanism, the reset mechanism, and the way the model may exhibit dynamic behaviours in phase space. It is shown experimentally in this paper that both the reset mechanism and the self-feed back control mechanism are important for the NDS model to work and to stabilise to one of the large number of available unstable periodic orbits (UPOs) that are embedded in its attractor. The experimental investigation suggests that the internal dynamics of the NDS neuron provide a rich set of dynamic behaviours that can be controlled and stabilised. These wide range of dynamic behaviours may be exploited to carry out information processing tasks.
\end{abstract}

\section{Introduction}
\label{sec:intro}
It has been suggested by biologists that chaos may be one of the main ingredients in carrying out information processing tasks in human brain\cite{Babloyantz1996,Bershadskii2011,Crook2008,Destexhe1994,Frank1990,Scarda1987,Freeman1985,Freeman2000,Freeman1994b,Theiler1995,Wu2009}. Plugging chaos into Artificial Neural Networks (ANN) may enrich its state space with a large number of dynamic behaviors when compared to hopfield networks. These behaviors can be utilized through chaos control methods\cite{Ott1990,Pasemann1998,Pyragas1992}. Results of applying such control methods to chaotic attractors showed stabilization into one of its UPOs \cite{Crook2005,Crook2003,Fourati2010,Piccirillo2009,Wang2009}.

Chaotic neural network models have been devised to exploit the rich dynamic behaviours that chaos provides. The main idea is to apply appropriate control methods to map a specific input to one of the internal dynamic behaviours that the chaotic attractor encompasses. One way to do so is to use spikes as input signals. One system that have implemented the aforementioned idea is the NDS model \cite{Crook2005}. The model has been proposed initially in \cite{Crook2003} where the authors defined a chaotic neuron that is based on a discrete version of R\"{o}ssler system \cite{Rossler1976}. The idea is to provide a large number of memories using only one NDS neuron. A modified version of Pyragas \cite{Pyragas1992} has been used to control the dynamics of the NDS model. The authors in \cite{Crook2005} have shown that it is possible to retrieve previously stored periodic patterns using the modified control mechanism.

The authors in \cite{Goh2007} have used Lorenz attractor instead of R\"{o}ssler. They have used the idea of transient computation machine to detect human motion from sequences of video frames. In another research paper, it has been proposed that chaos may equip mammalian brain with the equivalent of kernel trick for solving hard nonlinear problems\cite{Crook2008} using the NDS neuron model. Spike Time Dependent Plasticity (STDP) has been studied in networks of NDS neurons \cite{Aoun2010}. The results of experiments suggest that the NDS neuron may own the realism of biological neural networks.

Even though the NDS neuron model has many interesting properties, it has some drawbacks including: the way the control and reset mechanisms work are unclear, the analysis of the dynamics of the model is missing, analysis of the stability of networks of such neurons is also missing, and learning algorithms need to be developed to exploit the large number of UPOs that the attractor of the NDS encompasses. In this paper the control and the reset mechanisms are investigated experimentally. This investigation is important to understand how the control mechanism works, and will show how important the reset mechanism is for the model to work.

The paper is organized as follows: section~\ref{sec:nds_model} describes the NDS model, section~\ref{sec:reconstruct} is devoted to investigate reconstructing the internal dynamics in NDS model, where section~\ref{subsec:control} investigates the NDS control mechanism and section~\ref{sec:nds_reset_inv} investigates the NDS reset mechanism.  In section~\ref{sec:discuss} the results of the experiments are discussed and finally section~\ref{sec:conclusion} concludes the paper.

\section{Describing the NDS model}
\label{sec:nds_model}
Crook et al. have proposed a chaotic spiking neuron model \cite{Crook2005}, viz., the NDS neuron. The NDS neuron is a conceptual discretized model that is based on a modified version of R\"{o}ssler's chaotic system \cite{Rossler1976}.

\begin{figure}[h]
\centering
\includegraphics[scale=0.6]{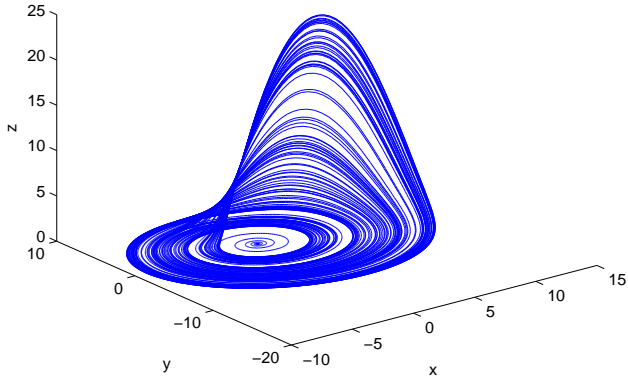}
\caption{R\"{o}ssler chaotic attractor with parameters $a= 0.2$, $b=0.2$, and $c=5.7$.}
\label{fig:FIG01}
\end{figure}

R\"{o}ssler system \cite{Rossler1976} is a simple dynamical system that exhibits chaos and has only one nonlinear term in its equations. R\"{o}ssler built the system in 1976, and it describes chemical fluctuation and is represented by the following differential equations:

\begin{equation}
\label{eqn:rslr_x} x' = -y-z
\end{equation}

\begin{equation}
\label{eqn:rslr_y} y' =x+a y
\end{equation}

\begin{equation}
\label{eqn:rslr_z} z' =b+z(x-c)
\end{equation}

Where $a$ and $b$ are usually fixed and $c$ is varied. The familiar parameter settings for R\"{o}ssler attractor are  $a= 0.2$, $b=0.2$, and $c=5.7$, and the corresponding attractor is shown in figure~\ref{fig:FIG01}.

The NDS model is a discrete version of R\"{o}ssler system. The main reason for discretization is that spikes should occur in discrete time. The discretization has been carried out by scaling the system variables ${x(t), y(t)}$ and  ${u(t)}$ using different scaling constants: $b,c$ and $d$. The values of these constants have been tuned experimentally until the dynamics of R\"{o}ssler system are preserved. If the values of these constants are larger, then a system trajectory will miss many dynamic evolutions while moving from one discrete iteration to the next.

The NDS model simulates a novel chaotic spiking neuron and is represented by:

\begin{equation}
\label{eqn:nds_x}
x(t+1) =x(t)+b(-y(t)-u(t))
\end{equation}

\begin{equation}
\label{eqn:nds_y}
y(t+1) = y(t)+c(x(t)+ay(t))
\end{equation}

\footnotesize
\begin{equation}
\label{eqn:nds_u}
u(t+1) =
\left\{
\begin{array}{ll}
\eta\sb{0} & u(t)>\theta\\
u(t)+d(v-u(t)x(t)+ku(t))+D(t) & u(t)\leq \theta
\end{array}
\right.
\end{equation}
\normalsize

\begin{equation}
\label{eqn:nds_D}
D(t) = F(t)+I(t)
\end{equation}

\begin{equation}
\label{eqn:nds_Fb}
F(t) = \sum_{j=1}^{n} w_j \gamma(t-\tau_j)
\end{equation}

\begin{equation}
\label{eqn:nds_In}
I(t) = \sum_{j=1}^{n}I_j(t)
\end{equation}

\begin{equation}
\label{eqn:nds_gamma}
\gamma(t+1) =
\left\{
\begin{array}{ll}
1 & u(t)>\theta\\
0 & u(t)\leq \theta
\end{array}
\right.
\end{equation}

\normalsize

where ${x(t), y(t)}$ and  ${u(t)}$ describe the internal state of the neuron, ${\gamma(t)}$  is
the neuron's binary output, $F(t)$ represent the feedback signals, $In(t)$ is the external input binary spike train, and the constants and parameters of the model are: ${a=v=0.002}$, ${b = c = 0.03}$, ${d = 0.8}$, ${k = -0.057}$, ${\theta = -0.01}$, ${\eta\sb{0}=-1}$ and $\tau\sb{j}$ is the period length of the feedback signals.

The term ${\sum_{j=1}^{n} w_j\gamma(t-\tau_j)}$ in equation~\ref{eqn:nds_Fb} represents the self-feedback mechanism which is the summation of spikes fired from the NDS neuron at time ${t=t-\tau\sb{j}}$ each multiplied by the corresponding connection weight $w\sb{j}$.

\begin{figure}[ht]
\centerline{\includegraphics[width=3.5in]{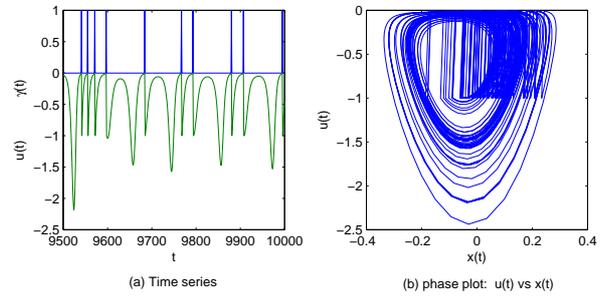}}
\caption{The chaotic behaviour of a NDS neuron without input (a) the time series of $u(t)$ and $\gamma(t)$, and (b) the phase space of $x(t)$ versus $u(t)$}
\label{fig:FIG02}
\end{figure}

By varying the system parameters such as period length ${\tau}$, connection time delays and initial conditions, large number of distinct orbits with varying periodicity may be stabilised. The dynamics of a single NDS neuron without input is shown in Figure~\ref{fig:FIG02}.

The NDS model was built as a spiking neural model, therefore R\"{o}ssler system equations have been modified to cope with the reset mechanism after spikes are emitted. The modification that was made to R\"{o}ssler system was the change of the sign of the term ${u(t)x(t)}$ in equation~\ref{eqn:rslr_z} from positive to negative ($u$ here is equivalent to $z$), the term then becomes ${-u(t)x(t)}$. This modification will flip the fin of R\"{o}ssler attractor in the ${x-u}$ plane horizontally. This effect can be noticed once figures~\ref{fig:FIG02} and ~\ref{fig:FIG01} are compared.

Before the control and reset mechanisms are investigated; the method of reconstructing the internal dynamics is explained first.

\section{Reconstructing internal dynamics in NDS}
\label{sec:reconstruct}
\begin{figure}[ht]
\centerline{\includegraphics[scale=0.6]{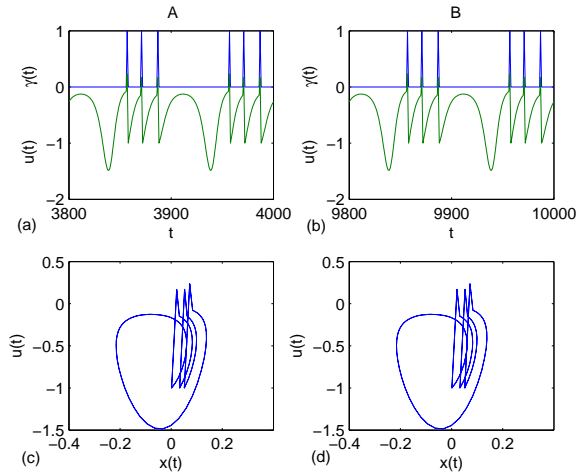}}
\caption{The reconstruction of period-3 orbit with time delay 100.
(a) and (c) show the behaviour of the feedback controlled neuron.
(b) and (d) show the behaviour of the reconstructed period with forcing inputs.}
\label{fig:FIG04}
\end{figure}

One of the main properties of the NDS model is the ability of reconstructing the original periodic internal dynamics of the neuron. Experimental results have shown this to be possible by stimulating an NDS neuron with a previously recorded firing pattern of a stabilised orbit \cite{Crook2007a}; this forces the chaotic dynamics of the neuron into the same orbit. Even though the mechanism of the stabilization, namely periodic forcing \cite{Schuster1998}, is completely different from the feedback control, the orbit is the same in both cases. An example is shown in Figure~\ref{fig:FIG04} where (a) and (c) are the original dynamics, and (b) and (d) are the successfully reconstructed dynamics which can be seen to be identical.

\begin{figure}[ht]
\centerline{\includegraphics[scale=0.6]{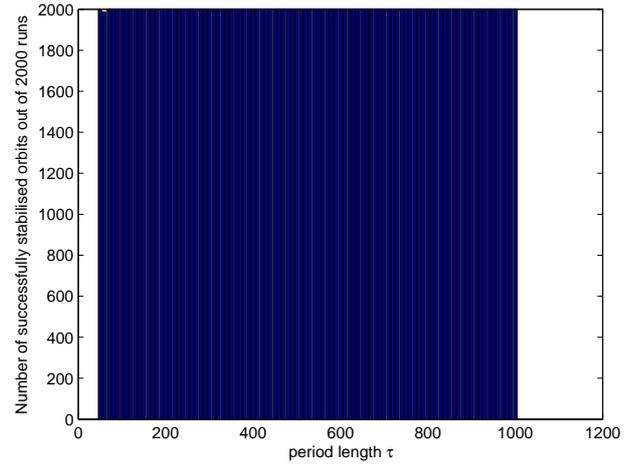}}
\caption{The number of successfully stabilised UPOs for different periodicity ${\tau}$ lengths of a NDS neuron with time-delayed
self-feedback connection.}
\label{fig:FIG05}
\end{figure}

Figure~\ref{fig:FIG04} shows just one example of the stabilised orbits, however there are very large number of stabilised orbits for different period lengths ${\tau}$. To demonstrate this, an experimentation has been set up to calculate the reliability of the NDS model in stabilizing orbits. A single NDS neuron with a time-delayed self-feedback connection is used with different period lengths ${\tau \in}$ ${[50,1000]}$. For each ${\tau}$ $2000$ different initial conditions have been used, and the system then run and the time series of the system variables are recorded. Figure~\ref{fig:FIG05} shows how reliable is the stabilization of orbits for different periods of length ${\tau}$ using forcing. Also the stabilization reliability was calculated and it is  ${99.9922\%}$.

The control mechanism is a very important aspect of the NDS model and will be investigated in some detail in the following section.

\section{NDS Control Investigation}
\label{subsec:control}
In the following set of experiments the control mechanism
that is used in the NDS model will be investigated in some detail.
The purpose of these experimentations is to understand how the
internal dynamics respond to either the self-feedback signals or
external input signals. There will be two experimental setups:

\begin{itemize}
    \item Comparing the variables of the NDS model.
    \item Replacing the feedback signal with a fixed value.
\end{itemize}

\subsection{Comparing the variables of the NDS model}
\label{sub_sec:nds_var_cmpr}
This set of experiments investigates the control mechanism in the case of applying time-delayed self-feedback signals in terms of the internal dynamics of the NDS. The idea is to compare two sets of points ${\rho(t)}$ and ${\rho(t-\tau)}$ where ${\rho \in [x,y,u]}$.

The results will be considered for each variable separately and then also for all three state variables $x,y$ and $u$ combined in 3D phase space. In all of the experimentations the self-feedback control mechanism is applied at time step ${t=1001}$. This may help in eliminating any transient behaviour.

In this experiment setup one NDS neuron is used with a self feedback connection with time delay ${\tau =100}$ and connection weight of $0.3$ which is initially turned off. The other parameters values are the same as described in equations~\ref{eqn:nds_x},~\ref{eqn:nds_y} and ~\ref{eqn:nds_u}. The internal dynamics evolves until time step ${t=1000}$ when the self-feedback connection is turned on and the internal dynamics continue evolving until time step ${t=5000}$. The time series of all the variables and output spikes are recorded.

\begin{figure}[ht]
\centerline{\includegraphics[scale=0.6]{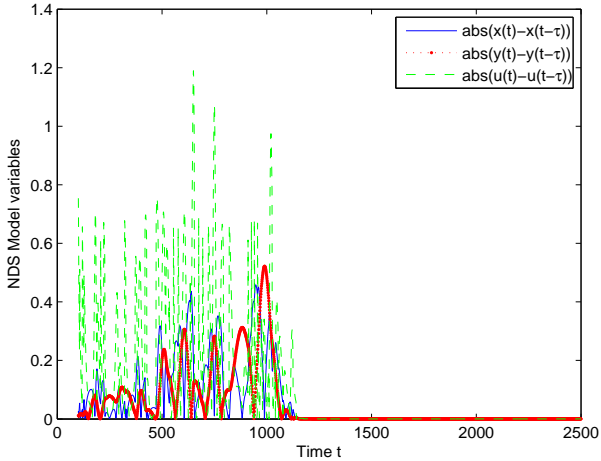}}
\caption{Comparing internal variable values of ${x,y}$ and ${u}$
at time step ${t}$ with the same variables values at time step ${t-\tau}$}
\label{fig:FIG03}
\end{figure}

Figure~\ref{fig:FIG03} depicts the absolute difference between the time series of the variables ${x(t),y(t)}$ and ${u(t)}$ along ${x(t-\tau), y(t-\tau)}$ and $u(t-\tau)$. Note from the figure that once the self-feedback connection is turned on the values of ${x,y}$ and ${u}$ variables at time step ${t}$ tend to synchronize with the values at time step ${t-\tau}$ and a corresponding spike pattern emerges. Note also how the absolute difference represents a nonlinear decreasing pattern (which will be used later as a \emph{'difference pattern'}) which reflects the dynamics encapsulated within the variables. For variable ${x}$ the dynamics is the product of the nonlinear term $-u\sb{i}(t)x\sb{i}(t)$ of equation~\ref{eqn:nds_u}.However, for variable $y$ the absolute difference is not as nonlinear as the one shown with the ${x}$ variable and is more likely to be linear pattern. This is because the variable $y$ appears in a linear form in the NDS system equations. It is important here to mention that the behaviour that appears in figure~\ref{fig:FIG03} is not an average and is the result of one run that starts from a random initial condition.

Finally, variable ${u}$ has more nonlinearity in the difference pattern when compared with ${x}$ difference pattern. This difference in nonlinearity is due to the nonlinear term that is found in the ${u}$ variable equation which is the only variable being updated based on a nonlinear term. Even though the nonlinear term affects the ${x}$ variable value, as the ${u}$ variable is included in the equation of the ${x}$ variable, still the dynamics are not as strong as appears in the ${u}$ variable values as in the difference pattern mentioned previously.

\begin{figure}[ht]
\centerline{\includegraphics[scale=0.6]{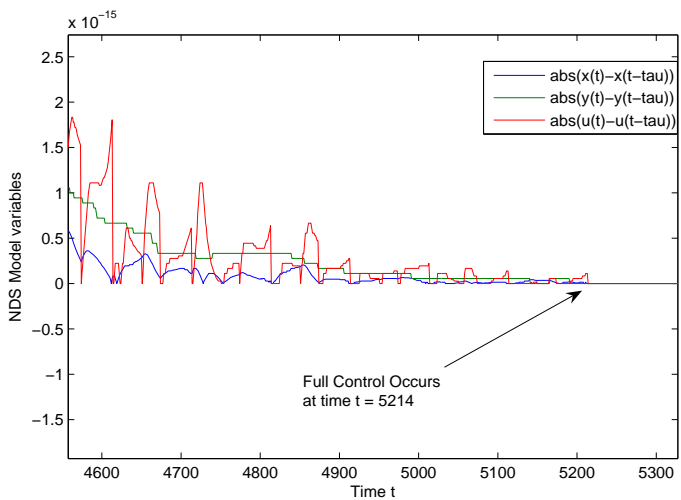}}
\caption{Full dynamic control while comparing internal variable values
of ${x,y}$ and ${u}$ at time step ${t}$ with the same variables values
at time step ${t-\tau}$}
\label{fig:FIG06}
\end{figure}

One observation in this experimentation setup is the high sensitivity of the model. On average, the spike output is stabilised by around ${t=1385}$. However, the underlying dynamics often do not stabilize until much later (average of ${t=5000}$). One example is shown in figure~\ref{fig:FIG06}. This demonstrates the dynamical behaviour of the system and may be due to the nonlinearity expressed by the $u$ variable.

\begin{figure}[ht]
\centerline{\includegraphics[scale=0.6]{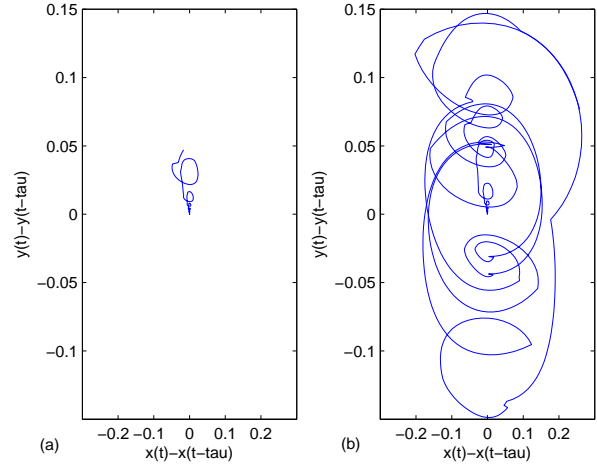}}
\caption{Comparing internal variable values of ${x}$ and ${y}$
at time step ${t}$ with the same variables values at time step
${t-\tau}$} \label{fig:FIG07}
\end{figure}

To see how the $x$ and $y$ variables behave together in phase space, the absolute difference between these variables time series at current time and the previous period of time is depicted in figure~\ref{fig:FIG07}.

 Figure ~\ref{fig:FIG07} shows how the difference of the variables values together between two successive periods of times of length ${\tau}$ spirals around in circles until it reaches ${0}$. Note that moving from one circle to the other looks like a transient which is mainly the effect of a neuron firing a spike. These are obviously seen in the figure where the difference trajectory is not smooth.

\begin{figure}[ht]
\centerline{\includegraphics[scale=0.6]{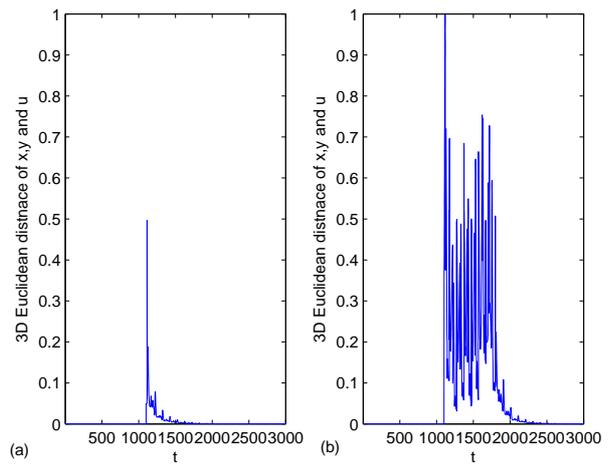}}
\caption{The Euclidean distance in three dimensions of ${x,y}$ and ${u}$ at time step ${t}$ with the same variables values at time step ${t-\tau}$}
\label{fig:FIG08}
\end{figure}

In figure~\ref{fig:FIG08} the Euclidean distance in three dimensions, between the values of ${x,y}$ and ${u}$ variables at time step ${t}$ and those at time step ${t-\tau}$, is calculated and depicted in order to investigate the way the system internal dynamics synchronize to itself. The figure shows that the difference pattern is an average of the dynamic behaviour of the three variables together, therefore the difference pattern seems to have less dynamical behaviour compared to those shown in figure~\ref{fig:FIG06}.

These experimental results which are shown in figures~\ref{fig:FIG07}-~\ref{fig:FIG08}
help in understanding how the control mechanism might be working and might also address the importance of the self-feedback signal.

The control mechanism seems to be a self~-~synchronization process. This appears from Figures~\ref{fig:FIG07}-~\ref{fig:FIG08} where the time series of one period of time ${\tau}$ of the system variables is coming closer to the stabilised orbit each ${\tau}$ period of time. The reason behind this self-synchronization is the time-delayed self-feedback signal that encourages the $u$ variable value to cross the threshold, even though it is not guaranteed to happen every ${\tau}$ period of time and depends on the value of the $u$ variable just before receiving the signal. If the value of the $u$ variable crosses the threshold ${\theta}$ then the reset occurs and the $u$ variable is assigned the value of ${\eta\sb{0}}$. Because the time-delayed self-feedback will incur a signal every ${\tau}$ period of time, then the value of the $u$ variable will have a periodic value when reset occurs at specific times. These specific times will organize the time-series of period ${\tau}$ and the value of the $u$ variable at these time steps will affect the values of both $x$ and $y$ variables, and hence the internal system dynamics are driven to synchronize with itself and stabilize to one of the attractor UPOs.

The results also show the strength of the dynamics of individual variables and suggest that: ${y}$ variable has a weak dynamics, ${x}$ variable has an intermediate dynamics, and ${u}$ variable has the strongest dynamics amongst the three. At the same time the average of these dynamics is shown in figure~\ref{fig:FIG08} where the dynamics of the difference pattern has less \textbf{rapidly changing dynamics} than that of the ${u}$ variable and more dynamics than the ${y}$ variable has.

\subsection{Replacing the feedback signal with a fixed value}
\label{sub_sec:nds_fix_cmpr}
The aim of this experimental setup is to investigate the importance of timing of time-delayed self-feedback connection compared to the connection weight in the process of controlling the NDS internal dynamics.
One method to investigate this is by replacing the self-feedback signal with a fixed value to be assigned to the ${u}$ variable periodically. Then the same experimental set up that has been used previously will be used with one change: replace the time-delayed self-feedback connection with a fixed value to be assigned to the ${u}$ variable in a specific time every ${\tau}$ period of time.

In one example of this experimental setup the fixed value is chosen to be ${1}$, which will be assigned to the ${u}$ variable at a fixed time of ${t=3}$ every ${\tau}$ period of time as an input. Initially, the internal dynamics evolves from a randomly chosen initial point until ${t=1000}$ without input. After that, the fixed value is added to the variable ${u}$ values at the times like ${t=1003,1103,1203,... etc}$ until ${t=30000}$. Then the time series of all the variables and output spikes are recorded.

\begin{figure}[ht]
\centerline{\includegraphics[scale=0.6]{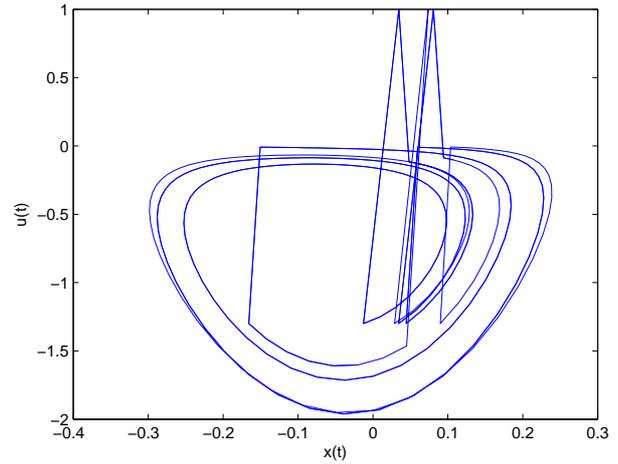}}
\caption{${x(t)}$ vs. ${u(t)}$  stabilised orbit using fixed value
of ${1}$ being assigned to u variable rather than the self-feedback
connection value.}
\label{fig:FIG09}
\end{figure}

\begin{figure}[ht]
\centerline{\includegraphics[scale=0.6]{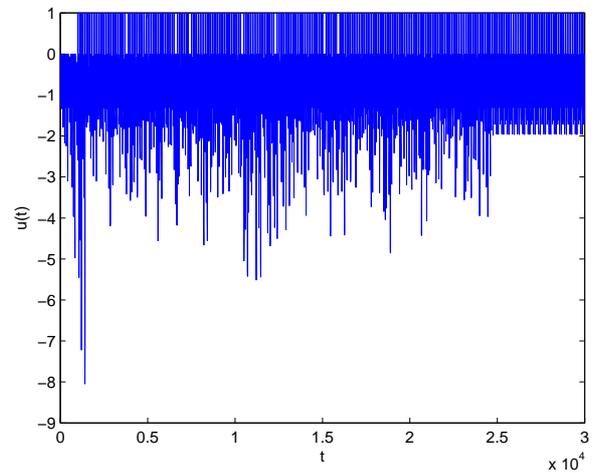}}
\caption{${u(t)}$ time series using fixed value of ${1}$ being
added to ${u}$ variable rather than the self-feedback connection
value.} \label{fig:FIG10}
\end{figure}

The results of this experimentation have shown that UPOs can be stabilised, and one of these UPOs is depicted in figure~\ref{fig:FIG09}. In one run, the time series for ${x,y}$ and ${u}$ variables have been recorded and have been depicted for variable ${u}$ as shown in figure~\ref{fig:FIG10}. The figure shows that it takes the internal dynamics a very long time compared to the time it takes an UPO to be stabilised when using time-delayed self-feedback connection.

The spike train for the stabilised orbit shown in Figure~\ref{fig:FIG09} is (3,84,103,194, 203,286,303,384,403,494) and is of period length ${5\tau}$. Note that the periodicity of the stabilised orbit has increased dramatically compared to UPOs with periodicity ${\tau}$ when self-feedback connection is used. This happened because the external input will not drive the internal dynamics to synchronize with itself every $\tau$ time steps. This increase in periodicity occurs despite the fact that the fixed value signal is fed every ${\tau}$ time steps. The spike train mentioned above is fed to an NDS neuron that is equipped with a time-delayed self-feedback connection when  ${t\in([1000,1500]}$ and an UPO was stabilised accordingly which has a periodicity of ${\tau}$ and with spike train of ${(3,38,62,82)}$ for one chosen initial condition. The difference in periodicity of the two stabilised UPOs is due to the time-delayed self-feedback connection. In the first case without the self-feedback connection the internal dynamics are not restricted to the periodicity ${\tau}$, because the output spikes of the system are not periodic in the sense that they are not fed back to the model, therefore the output spikes every ${\tau}$ period of time are different from those in the previous ${\tau}$ period of time, and consequently it took a longer time to stabilize to a UPO. On the other hand, when the model is equipped with a self-feedback connection, the output spikes are fed back to the model by means of the connection and hence encourages the internal dynamics to repeat every ${\tau}$ of time.

Even though UPOs can still be stabilised when no connection is present, it takes a very long time to stabilize a UPO, and those stabilised UPOs cannot be reconstructed by feeding the spike train to an NDS neuron with a self-feedback connection. Moreover, in the case of feeding the spike train $(3,84,103,194,203,286,$ $303,384,403,494)$ to an NDS model without a self~-~feedback connection, there was no stabilised orbit even when the input spike train is being fed to the neuron model during all the running time.

Based on the previous results, a UPO of an NDS model can be defined as the orbit that has output spikes occurring within a periodicity of ${\tau}$, which can be reconstructed by feeding these spikes to an NDS model with or without a self-feedback connection with time delay ${\tau}$.

The previous discussions substantiate that the self-feedback signal is an important ingredient for the NDS model to be stabilised to one of the large number of UPOs that the attractor encompasses. Also the self-feedback control is powerful in stabilizing orbits in terms of time; because the firing times are being adapted to a specific pattern, which means the spike train will converge to one of the available UPOs.

\section{Reset mechanism investigation}
\label{sec:nds_reset_inv}
In this section the reset mechanism of the NDS neuron model is investigated to understand its role in stabilization. Investigation includes studying the NDS stabilization process in the case of replacing the reset fixed value by a relative reset value. Also, finding the range for the reset value will help in utilizing the NDS attractor. In the coming two subsections the aforementioned investigations is explained.

\subsection{Replace the reset fixed value by a relative reset value}
\label{sub_sec:nds_relative_reset}
The reset mechanism is a crucial element of the NDS model because without it the internal dynamics will evolve to infinity. However to investigate the role of the reset mechanism in stabilizing UPOs then the model will be studied with a relative reset value rather than an absolute value.

The same setup used previously is used here with the exception of replacing the fixed reset value ${\eta\sb{0}}$ with a relative reset mechanism. The idea is to add the reset value ${\eta\sb{0}}$ to the value of the ${u}$ variable instead of assigning ${\eta\sb{0}}$ to the ${u}$ variable. This mechanism will allow the $u$ variable to have different positions when reset occurs until a UPO is stabilised. Then the reset positions will be a periodic pattern and each position is assigned with an output spike.

\begin{figure}[ht]
\centerline{\includegraphics[scale=0.6]{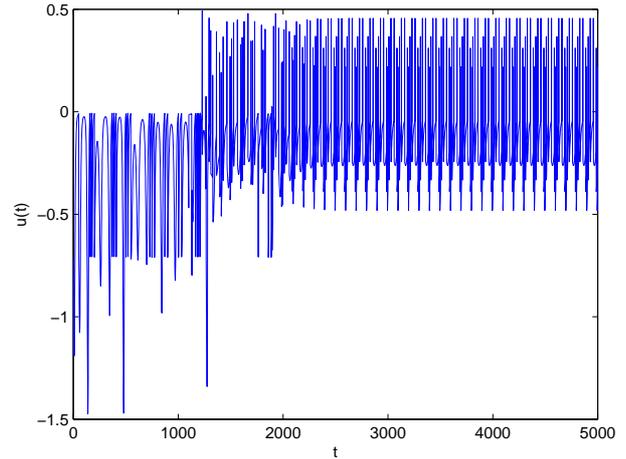}}
\caption{${u(t)}$ time series with relative reset value is used rather than fixed reset value.} \label{fig:FIG11}
\end{figure}

The results of one experimental run are shown in figure~\ref{fig:FIG11}. Note that it takes on average ${350}$ time steps for the spike output to be stabilised. This is a little lower than the average when using fixed reset value, and the reason behind this is that the values of the ${u}$ variable after reset are a little greater than those when using fixed reset value, therefore the time the value of the ${u}$ variable needs to reach threshold is shorter, and hence the firing rate is greater. However, the average time for the internal dynamics to be fully stabilised when using fixed reset is ${4035}$ which is lower than when compared to when using relative reset mechanism (${4085}$). This might be the result of the different positions of the values of the ${u}$ variable for every reset before stabilization occurs, which affects the speed in which the ${u}$ variable drives the variables ${x}$ and ${y}$.

The previous discussions suggest that the reset mechanism is working because the value of the $u$ variable is being reset to a specific area in the phase space. However, to investigate the previous conclusion; the next experimentation setup will show the range of values that can be used as a reset value for the NDS model with a fixed reset mechanism.

\subsection{Finding the range for the reset value}
\label{sub_sec:nds_reset_range}
The same experimental setup used previously is used here except that the reset value ${\eta\sb{0}}$ will be adjusted to different values. These values are chosen to be in the range ${\eta\sb{0}\in[-2,0.1]}$ for threshold ${\theta=-0.01}$, this range represents values above threshold, around threshold, and below threshold. Therefore results of experiments are divided into three regions as the following paragraphs demonstrate.

\begin{figure}[ht]
\centerline{\includegraphics[scale=0.6]{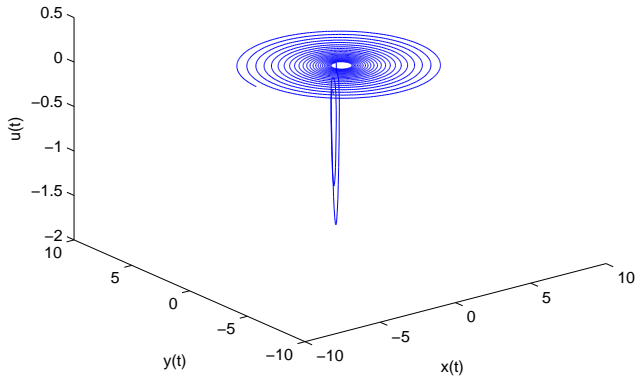}}
\caption{NDS attractor for reset value ${\eta\sb{0}=0.1}$}
\label{fig:FIG12}
\end{figure}

The results for runs performed with the reset value being assigned to a value above threshold show that the system internal dynamics is reduced to two dimensions composed of ${x}$ and ${y}$ variable. The behaviour of the trajectory is a spiral repellor around point ${(0,0)}$ in the ${x-y}$ plane. The results of one the runs are shown in figure~\ref{fig:FIG12} where the reset value was chosen to be ${0.1}$. Note that the trajectory goes in a transient first until it reaches the threshold value, and then it spirals around the origin. The reason for this behaviour is that the ${u}$ variable value is always above threshold and is not changing, therefore the only changing variables are ${x}$ and ${y}$ which represent a simple oscillating system.

\begin{figure}[ht]
\centerline{\includegraphics[scale=0.6]{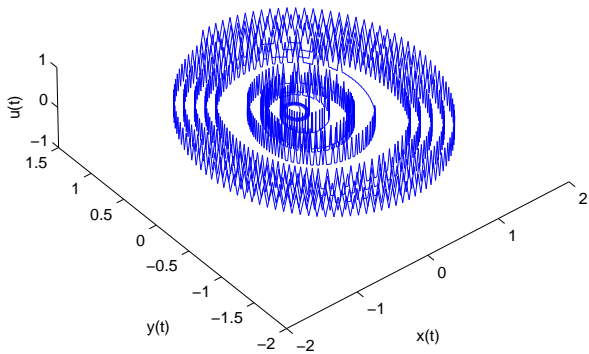}}
\caption{NDS attractor for reset value ${\eta\sb{0} = -0.02}$}
\label{fig:FIG13}
\end{figure}

The results for runs performed with the reset value being assigned to a value around and less than the threshold in the range of ${\eta\sb{0}\in[\theta-0.035,\theta]}$ show that the system internal dynamics is still linear but it appears in three dimensions. First the trajectory goes into a transient, and then spirals around the origin in the ${x-y}$ plane with the ${u}$ variable alternating between two values. The results of one run is shown in figure~\ref{fig:FIG13} where the reset value chosen to be ${-0.02}$. The reason for this behaviour is that the ${u}$ variable value is always very near to the threshold and it takes only a few time steps until the ${u}$ value reaches threshold value, therefore the variables ${x}$ and ${y}$ are dominating the system behaviour and again represent a simple oscillating system.

Finally, the third case is when the reset value is defined to be in the range ${\eta\sb{0}\in(-2,\theta-0.035]}$. Many experiment runs have been carried out with the reset value changed from ${\eta\sb{0}=-0.05}$ to ${\eta\sb{0}=-1.2}$. The results for these values show that the NDS attractor is reliable with percentage of around ${97\%}$. \textbf{Reliable} here means stabilizing UPOs quickly and more important keeping the system working and preventing it from approaching infinity. So a reliability of ${97\%}$ means that out of ${100}$ runs there were ${97}$ successfully stabilised orbits, and in ${3}$ times the system trajectories approach infinity. This happens because if the reset value is a deep negative number, then chances for the trajectory to leave the attractor are increased because if the $x$ variable exceeds a negative limit value then it will approach infinity.

\begin{figure}[ht]
\centerline{\includegraphics[scale=0.6]{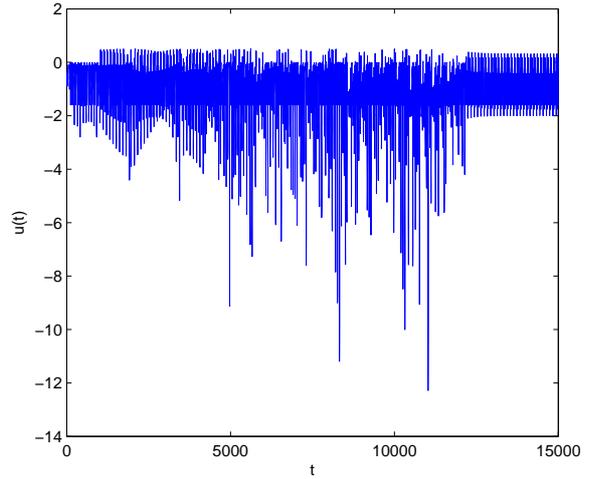}}
\caption{${u}$ variable time series for reset value ${\eta\sb{0}=-1.59}$ and initial conditions of (-0.1556,0.4469,-0.3596)}
\label{fig:FIG14}
\end{figure}

However, when the reset value changed to ${-1.4}$ the reliability percentage was reduced. The reset value continued to be changed until ${-2}$ where the reliability percentage becomes around ${5\%}$. Also it is important here to mention that the time it takes for a UPO to be stabilised increases when decreasing the reset value. In one experiment run the reset value was chosen to be ${\eta\sb{0}=-1.59}$ and the trajectory stabilised to one UPO after around $12500$ time steps as shown in Figure~\ref{fig:FIG14} where the time series of the ${u}$ variable values are depicted.

To summarise, optimal control is reliably achieved for values of ${\eta\sb{0}}$ in the range ${[-0.05,-1.2]}$. This allows accessibility to rich dynamics with high reliability.

\section{Discussion}
\label{sec:discuss}
The NDS model has been developed for the purpose of exploiting the rich dynamics that its attract provides in carrying out information processing tasks. This exploitation can be maximized if the way these dynamic behaviours that are exhibited in phase space is known. This has been carried out by experimental investigations of the internal dynamics of the NDS model. The results of these investigations partly explain how UPOs are stabilised in phase space and suggest optimal parameter settings to enlarge the capacity and ensure stability in NDS neurons.

The results of all the previous experiments suggest that both the reset mechanism and the feedback signal are the major ingredients for the NDS model to work and to be stabilised to one of its available UPOs.

Note from figure~\ref{fig:FIG09} that even though the system is fed with single signal each ${\tau}$ time steps, the system has still been stabilised to one of its UPOs. This single signal represent a special case of the time-delayed self-feedback mechanism with the exception of ignoring all the output spikes and only repeating one signal. This means that the feedback control is very important element in the NDS model.

Also note from figure~\ref{fig:FIG14} that the reset mechanism has led the internal dynamics to be stabilised after ${125\tau}$ time steps even though the reset value has increased to ${-1.59}$ which is almost double the original value defined for the NDS model (${-1}$). The reset values range that the NDS model performance is high has been found to be in ${[-0.05,-1.2]}$.  The reset mechanism is crucial for the NDS model to work because without the reset mechanism the internal dynamics will approach infinity.

The question that might be raised here is why both of them are important for the NDS model to be stabilised? The other question is how the control mechanism leads the internal dynamics to synchronize with itself? The experiments that have been conducted so far provide some insight into these issues. However, an analysis of the NDS model from a mathematical and dynamical systems perspective may suggest answers to these questions.

\section{Conclusion}
\label{sec:conclusion}

One chaotic model, viz., the NDS model has been investigated in this paper. The capabilities of the NDS neuron are considerable in terms of UPOs. These UPOs can be stabilised using a feedback control mechanism. The NDS model is a modified version of R\"{o}ssler chaotic system. The rich dynamics of R\"{o}ssler system is inherited by the NDS model. This is shown by the large number of UPOs that can be stabilised, and also is obvious from the NDS model attractor as shown in figure~\ref{fig:FIG02}.

First, the NDS model is described and its dynamics has been depicted and explained. This explanation gives an initial understanding for the behaviour of the NDS model and suggests how the model works.

Different experimental setups have been prepared to understand the dynamic behaviour of the NDS model in terms of control and reset mechanisms. The results of these experimentations have revealed that both the reset mechanism and the self-feed back control mechanism are important for the NDS model to work and to stabilise to one of the large number of UPOs that are embedded in the attractor. This wide range of dynamic behaviours may be exploited to carry out information processing tasks.

However these conclusions are based only on experimental results, therefore a mathematical analysis of the NDS model is needed to shed some light on these properties of the model, which will also help in understanding the different dynamical behaviours that the NDS model exhibits in phase space. This is being prepared and will be published soon.

\section{acknowledgements}
This work is a modified and updated part of a thesis submitted to the school of technology at Oxford Brookes University\cite{Alhawarat2007}. This work was supported by the University of Petra, Amman, Jordan. I'm very grateful to this support and would like to thank all the staff who were involved in helping and encouraging me.

\bibliographystyle{ijcaArticle}
\bibliography{MohdPapers}
\end{document}